# A Comparative Study of Gaussian Mixture Model and Radial Basis Function for Voice Recognition

Fatai Adesina Anifowose
Center for Petroleum and Minerals, The Research Institute
King Fahd University of Petroleum and Minerals
Dhahran 31261, Saudi Arabia
anifowo@kfupm.edu.sa

*Abstract*—A comparative study of the application of Gaussian Mixture Model (GMM) and Radial Basis Function (RBF) in biometric recognition of voice has been carried out and presented. The application of machine learning techniques to biometric authentication and recognition problems has gained a widespread acceptance. In this research, a GMM model was trained, using Expectation Maximization (EM) algorithm, on a dataset containing 10 classes of vowels and the model was used to predict the appropriate classes using a validation dataset. For experimental validity, the model was compared to the performance of two different versions of RBF model using the same learning and validation datasets. The results showed very close recognition accuracy between the GMM and the standard RBF model, but with GMM performing better than the standard RBF by less than 1% and the two models outperformed similar models reported in literature. The DTREG version of RBF outperformed the other two models by producing 94.8% recognition accuracy. In terms of recognition time, the standard RBF was found to be the fastest among the three models.

## I. INTRODUCTION

Biometrics is a measurable, physical characteristic or personal behavioral trait used to recognize the identity, or verify the claimed identity, of a candidate. Biometric recognition is a personal recognition system based on "who you are or what you do" as opposed to "what you know" (password) or "what you have" (ID card) [17]. The goal of voice recognition in biometrics is to verify an individual's identity based on his or her voice. Because voice is one of the most natural forms of communication, identifying people by voice has drawn the attention of lawyers, judges, investigators, law enforcement agencies and other practitioners of forensics.

Computer forensics is the application of science and engineering to the legal problem of digital evidence. It is a synthesis of science and law [8]. A high level of accuracy is required in critical systems such as online financial transactions, critical medical records, preventing benefit fraud, resetting passwords, and voice indexing.

In view of the importance of accurate classification of vowels in a voice recognition system, the need for a well-trained computational intelligence model with an acceptable percentage of classification accuracy (hence a low percentage of misclassification error) is highly desired. Gaussian Mixture Models (GMMs) and Radial Basis Function (RBF) networks have been identified in both practice and literature as two of the promising neural models for pattern classification.

The rest of this paper is organized as follows. Section II reviews the literature on voice recognition; overview and application of GMM and RBF in biometric voice recognition; and an overview of the RBF component of DTREG software. A description of the data and tools used in the design and implementation of this work are discussed in Section III. Section IV describes the experimental approach followed in this work and the criteria for quality measurement used to evaluate its validity. The results of the experiment are discussed in section V while conclusions are drawn in section VI.

## II. LITERATURE SURVEY

### A. Voice Recognition

A good deal of effort has been made in the recent past by researchers in their attempt to come up with computational intelligence models with an acceptable level of classification accuracy.

A novel suspect-adaptive technique for robust forensic speaker recognition using Maximum A-Posteori (MAP) estimation was presented by [1]. The technique addressed Likelihood Ratio computation in limited suspect speech data conditions obtaining good calibration performance and robustness by allowing the system to weigh the relevance of the suspect specificities depending on the amount of suspect data available via MAP estimation. The results showed that the proposed technique outperformed other previously proposed non-adaptive approaches.

[2] presented three mainstream approaches including Parallel Phone Recognition Language Modeling (PPRLM), Support Vector Machine (SVM) and the general Gaussian Mixture Models (GMMs). The experimental results showed that the SVM framework achieved an equal error rate (EER) of

4.0%, outperforming the state-of-art systems by more than 30% relative error reduction. Also, the performances of their proposed PPRLM and GMMs algorithms achieved an EER of 5.1% and 5.0% respectively.

Support Vector Machines (SVMs) were presented by [3] by introducing a sequence kernel used in language identification. Then a Gaussian Mixture Model was developed to do the sequence mapping task of a variable length sequence of vectors to a fixed dimensional space. Their results demonstrated that the new system yielded a performance superior to those of a GMM classifier and a Generalized Linear Discriminant Sequence (GLDS) Kernel.

Using a vowel detection algorithm, [4] segmented rhythmic units related to syllables by extracting parameters such as consonantal and vowel duration, and cluster complexity and modeled with a Gaussian Mixture. Results reached up to $86 \pm 6\%$ of correct discrimination between stress-timed, mora-timed and syllable-timed classes of languages. These were then compared with that of a standard acoustic Gaussian mixture modeling approach that yielded $88 \pm 5\%$ of correct identification.

[9] presented an additive and cumulative improvements over several innovative techniques that can be applied in a Parallel Phone Recognition followed by Language Modeling (PPRLM) system for language identification (LID), obtaining a 61.8% relative error reduction from the base system. They started from the application of a variable threshold in score computation with a 35% error reduction, then a random selection of sentences for the different sets and the use of silence models, then, compared the bias removal technique with up to 19% error reduction and a Gaussian classifier of up to 37% error reduction, then, included the acoustic score in the Gaussian classifier with 2% error reduction, increased the number of Gaussians to have a multiple-Gaussian classifier with 14% error reduction and finally, included additional acoustic HMMs of the same language with success gaining 18% relative improvement.

### B. Gaussian Mixture Model (GMM)

From a clustering perspective, most biometric data cannot be adequately modeled by a single-cluster Gaussian model. However, they can often be accurately modeled via a Gaussian Mixture Model (GMM) i.e., data distribution can be expressed as a mixture of multiple normal distributions [7].

Basically, the Gaussian Mixture Model with k components is written as:

$$p(y|\mu_1, \ldots, \mu_k, s_1, \ldots, s_k, \pi_1, \ldots, \pi_k) = \sum_{j=1}^{k} \pi_j \mathcal{N}\left(\mu_j, s_j^{-1}\right) \qquad (1)$$

where $\mu_j$ are the means, $s_j$ the precisions (inverse variances), $\pi_j$ the mixing proportions (which must be positive and sum to one) and N is a (normalized) Gaussian with specified mean and variance. More details on the component parameters and their mathematical derivations can be found in [10-13, 25, 26].

[5] presented a generalized technique by using GMM and obtained an error of 17%. In another related work, [10] described two GMM-based approaches to language identification that use Shifted Delta Costar (SDC) feature vectors to achieve LID performance comparable to that of the best phone-based systems. The approaches included both acoustic scoring and a GMM tokenization system that is based on a variation of phonetic recognition and language modeling. The results showed significant improvement over the previously reported results.

A description of the major elements of MIT Lincoln Laboratory's Gaussian Mixture Model (GMM)-based speaker verification system built around the likelihood ratio test for verification, using simple but effective GMMs for likelihood functions, a Universal Background Model (UBM) for alternative speaker representation, and a form of Bayesian adaptation to derive speaker models from the UBM were presented by [6]. The results showed that the GMM-UBM system has proven to be very effective for speaker recognition tasks.

[12] evaluated the related problem of dialect identification using the GMMs with SDC features. Results showed that the use of the GMM techniques yields an average of 30% equal error rate for the dialects in one language used and about 13% equal error rate for the other one.

Other related works on GMM include [11, 13].

### C. Radial Basis Function (RBF)

A RBF Network, which is multilayer and feedforward, is often used for strict interpolation in multi-dimensional space. The term 'feedforward' means that the neurons are organized in the form of layers in a layered neural network. The basic architecture of a three-layered neural network is shown in Fig. 1.

A RBFN has three layers including input layer, hidden layer and output layer. The input layer is composed of input data. The hidden layer transforms the data from the input space to the hidden space using a non-linear function. The output layer, which is linear, yields the response of the network.

The argument of the activation function of each hidden unit in an RBFN computes the Euclidean distance between the input vector and the center of that unit. In the structure of RBFN, the input data X is an I-dimensional vector, which is transmitted to each hidden unit. The activation function of hidden units is symmetric in the input space, and the output of each hidden unit depends only on the radial distance between the input vector X and the center for the hidden unit. The output of each hidden unit, $h_j$, $j = 1, 2, \ldots, k$ is given by:

$$h_j(x) = \phi(\|x - c_j\|) \qquad (2)$$

Where $\| \ \|$ is the Euclidean Norm, $c_j$ is the center of the neuron in the hidden layer and $\Phi()$ is the activation function.

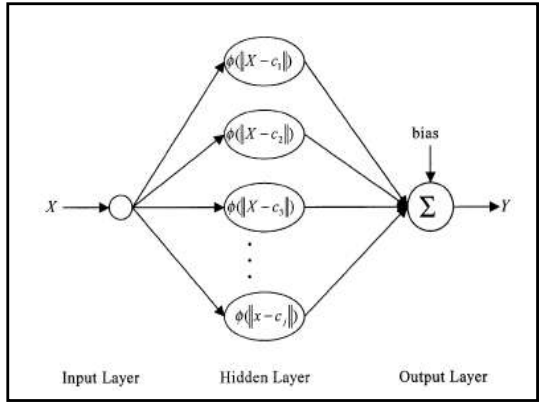

Figure 1.   The architecture of an RBF network.

The activation function is a non-linear function and is of many types   such Gaussian, multi-quadratic, thinspline and exponential functions. If the form of the basis function is selected in advance, then the trained RBFN will be closely related to the clustering quality of the training data towards the centers.

The Gaussian activation function can be written as:

$$\phi_j(x) = \exp\left[ - \frac{\|x - c_j\|^2}{2\rho^2} \right] \qquad (3)$$

where x is the training data and ρ is the width of the Gaussian function. A center and a width are associated with each hidden unit in the network. The weights connecting the hidden and output units are estimated using least mean square method. Finally, the response of each hidden unit is scaled by its connecting weights to the output units and then summed to produce the overall network output. Therefore, the k[th] output of the network $\hat{y}_k$ is:

$$\hat{y}_k = w_0 + \sum_{j=1}^{M} w_{jk}\phi_j(x) \qquad (4)$$

where $\Phi_j(x)$ is the response of the j[th] hidden unit, $w_{jk}$ is the connecting weight between the j[th] hidden unit and the k[th] output unit, and $w_0$ is the bias term [18].

RBF model, with its mathematical properties of interpolation and design matrices, is one of the promising neural models for pattern classification and has also gained popularity in voice recognition [14].

[15] presented a comparative study of the application of a minimal RBF Neural Network, the normal RBF and an elliptical RBF for speaker verification. The experimental results showed that the Minimal RBF outperforms the other techniques. Another work for explicitly modeling voice quality

variance in the acoustic models using RBF and Hidden Markov Models, in order to improve word recognition accuracy, was demonstrated by   [16]. They also presented SVM and concluded that voice quality can be classified using input features in speech recognition.

Other related works have been found in the fields of medicine [14], hydrology [18], computer security [19], petroleum engineering [20] and computer networking [21].

### D.   DTREG Radial Basis Function (DTREG-RBF)

DTREG software builds classification and regression Decision Trees, Neural and Radial Basis Function Networks, Support Vector Machine, Gene Expression programs, Discriminant Analysis and Logistic Regression models that describe data relationships and can be used to predict values for future observations. It also has full support for time series analysis. It analyzes data and generates a model showing how best to predict the values of the target variable based on values of the predictor variables. DTREG can create classical, single-tree models and also TreeBoost and Decision Tree Forest models consisting of ensembles of many trees. It includes a full Data Transformation Language (DTL) for transforming variables, creating new variables and selecting which rows to analyze [27].

One of the classification/regression tools available in DTREG is Radial Basis Function Networks. Like the standard RBF, DTREG-RBF has an input layer, a hidden layer and an output layer. The neurons in the hidden layer contain Gaussian transfer functions whose outputs are inversely proportional to the distance from the center of the neuron. Although the implementation is very different, RBF neural networks are conceptually similar to K-Nearest Neighbor (K-NN) models. The basic idea is that a predicted target value of an item is likely to be about the same as other items that have close values of the predictor variables.

DTREG uses a training algorithm developed by [28]. This algorithm uses an evolutionary approach to determine the optimal center points and spreads for each neuron. It also determines when to stop adding neurons to the network by monitoring the estimated leave-one-out (LOO) error and terminating when the LOO error begins to increase due to overfitting. The computation of the optimal weights between the neurons in the hidden layer and the summation layer is done using ridge regression. An iterative procedure developed by an author in 1966 is used to compute the optimal regularization lambda parameter that minimizes the Generalized Cross-Validation (GCV) error. A more detailed description of the DTREG can be found in [27].

### III.   DATA AND TOOLS

The training and testing data were obtained from an experimental 2-dimensional dataset available in [22]. The training data consists of 338 observations while the testing data consists of 333 observations. There are 2 input variables and each observation belongs to one of 10 classes of vowels to be classified using the trained models.

The GMM and RBF classifiers were implemented in MATLAB with the support of NETLAB toolbox obtained as freeware from [23] while the DTREG-RBF was implemented using the DTREG software version 8.2. The descriptive statistics of the training and test data are shown in table I and II while the scatter plots of the training and test data are shown in Fig. 2 respectively.

## IV. EXPERIMENTAL APPROACH AND CRITERIA FOR PERFORMANCE EVALUATION

The methodology in this work is based on the standard Pattern Recognition approach to classification problem using GMM and RBF. For training the models, Expectation Maximization (EM) algorithm was used for efficient optimization of the GMM parameters. The RBF used forward and backward propagation to optimize the parameters of the neurons using the popular Gaussian function as the transform function in the hidden layer as is common in literature. The parameters of the models were also tuned and varied and those with maximum classification accuracy were selected. The DTREG-RBF was run on the same dataset with the default parameter settings.

For the GMM, several runs were carried out using the "diag" and "full" covariance types and with number of centers ranging from 1 and 10 while for the RBF, several runs were carried out with different numbers of hidden neurons ranging from 1 and 36.

TABLE I. DESCRIPTIVE STATISTICS OF TRAINING data

| | X1 | X2 |
|---|---|---|
| Average | 567.82 | 1533.18 |
| Mode | 344.00 | 2684.00 |
| Median | 549.00 | 1319.50 |
| Std Dev | 209.83 | 673.94 |
| Max | 1138.00 | 3597.00 |
| Min | 210.00 | 557.00 |

TABLE II. DESCRIPTIVE STATISTICS OF TESTNING data

| | X1 | X2 |
|---|---|---|
| Average | 565.47 | 1540.38 |
| Mode | 542.00 | 2274.00 |
| Median | 542.00 | 1334.00 |
| Std Dev | 216.40 | 679.79 |
| Max | 1300.00 | 3369.00 |
| Min | 198.00 | 550.00 |

The DTREG-RBF is not flexible; only one variable can be set as the target at a time. It is most ideal for one-target classification problems. For this work, 10 different models were trained with each output column as the target. This was very cumbersome.

The most commonly used accuracy measure in classification tasks is the classification/recognition rate. This is calculated by:

$$\frac{p}{q} \ x \ 100$$

where $p$ is the number of correctly classified points and $q$ is the total number of data points.

For the purpose of evaluation in terms of speed of execution, Execution Time for training and testing was also used in this study.

## V. DISCUSSION OF RESULTS

For the GMM, generally, it was observed that the execution time increased as the number of centers was increased from 2, but with a little dip at 1. Similarly, the training and testing recognition rates increased as the number of centers was increased from 1 to 2 but decreased progressively when it was increased from 3. Fig. 3 and 4 show the plots of the different runs of the "diag" and "full" covariance types and how execution time and recognition rates vary with the number of centers. The class boundaries generated by the GMM Model for training and testing are shown in Fig. 5.

The results for GMM above showed that the average optimal performance was obtained with the combination of "full" covariance type and number of centers chosen to be 2.

For the RBF, generally, the training time increased as the number of hidden neurons increased while the testing time remained relatively constant except for little fluctuations. Also, the training and testing times increased gradually as the number of hidden neurons increased until up to 15 when they began to fall gradually at some points and remained relatively constant except for little fluctuations at some other points. Fig. 6 shows the decision boundaries of the RBF-based classifier using the same training and testing data applied on the GMMs while Fig. 9 shows the contour plot of the RBF model with the training data and the 15 centers.

The results for RBF above showed that the average optimal performance was obtained when the number of hidden neurons is set to 15.

As mentioned earlier in section IV, one disadvantage of the DTREG-RBF is that it accepts only one variable as the target. This constitutes a major restriction and poses a lot of difficulties. For each of the 10 vowel classes, one model was built by training it with the same dataset but with its respective class for classification. There is no automated way of doing this. For the purpose of effective comparison, the average of the number of neurons, training times and training and testing recognition rates were taken. Fig. 7 and 8 show the relationship between the number of hidden neurons and the execution time and classification accuracy respectively. They both indicate that the optimal performance in terms of execution time and classification accuracy is obtained approximately at the point where the number of hidden neurons is set to 15.

Comparatively, in terms of execution time, RBF clearly outperforms GMM and DTREG-RBF, but in terms of recognition rate, it was not clearly visible to see which is better between GMM and RBF since GMM (79.6%) is better in training than RBF (78.1%) while RBF (80.8%) is better in recognition than GMM (79.9%). To ensure fair judgment, the average of the training and testing recognition rates of the two models shows that GMM (79.7%) performs better than RBF (79.4%) by a margin of 0.3%. It is very clear that in terms of recognition accuracies, the DTREG-RBF model performed best with an average recognition rate of 94.79%. This is clearly shown in Fig. 10.

## VI. CONCLUSION

A comparative study of the application of Gaussian Mixture Model (GMM) and Radial Basis Function (RBF) Neural Networks with parameters optimized with EM algorithm and forward and backward propagation for biometric recognition of vowels have been implemented. At the end of the study, the two models produced 80% and 81% maximum recognition rates respectively. This is better than the 80% recognition rate of the GMM proposed by Jean-Luc et al. in [4] and very close to their acoustic GMM version with 83% recognition rate as well as the GMM proposed by [5]. The DTREG version of RBF produced a landmark 94.8% recognition rate outperforming the other two techniques and similar techniques earlier reported in literature.

This study has been carried out using a vowel dataset. The DTREG-RBF models were built with the default parameter settings left unchanged. This was done in order to establish a premise for valid comparison with other studies using the same tool. However, as at the time of this study, the author is not aware of any similar study implemented with the DTREG software, hence there is no ground for comparison with previous studies.

Further experimental studies to evaluate the classification and regression capability of DTREG will be carried out to use each of its component tools such as Support Vector Machines, Probabilistic and General Regression Neural Networks, Cascaded Correlation, Multilayer Perceptron, Decision Tree Forest, and Logistic Regression for various classification and prediction problems in comparison with their standard (usually MATLAB-implemented) versions.

Furthermore, in order to increase the confidence in this work and establish a better premise for valid comparison and generalization, a larger and more diverse dataset will be used. In order to overcome the limitation of the dataset used where a fixed data was preset for training and testing, we plan for a future study where stratified sampling approach will be used to divide the datasets into training and testing sets as this will give each row in the dataset an equal chance of being chosen for either training or testing each time the implementation is executed.

With our previous work on the hybridization of machine learning techniques [29], a study has commenced for the combination of GMM and RBF as a single hybrid model to achieve better learning and recognition rates. It has been reported [30-33] and confirmed [29] that hybrid techniques perform better than their individual components used separately.

### ACKNOWLEDGMENT

The author is grateful to the Department of Information and Computer Science and the College of Computer Sciences & Engineering of King Fahd University of Petroleum and Minerals for providing the computing environment and the licensed DTREG software for the purpose of this research. The supervision of Dr. Lahouari Ghouti and the technical evaluation Dr. Kanaan Faisal are also appreciated.

AUTHOR'S PROFILE

Fatai Adesina Anifowose was formerly a Research Assistant in the department of Information and Computer Science, King Fahd University of Petroleum and Minerals, Saudi Arabia. He now specializes in the application of Artificial Intelligence (AI) while working with the Center for Petroleum and Minerals at the Research Institute of the same university. He has been involved in various projects dealing with the prediction of porosity and permeability of oil and gas reservoirs using various AI techniques. He is recently interested in the hybridization of AI techniques for better performance.

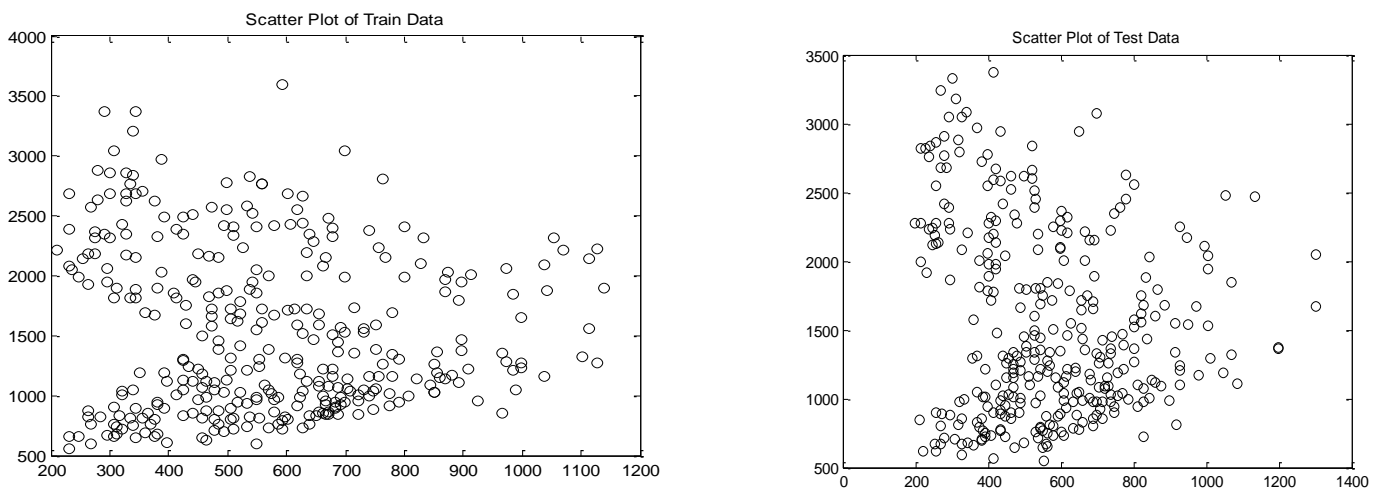

Figure 2. Scatter plot of training data with 338 observations and test data with 333 observations.

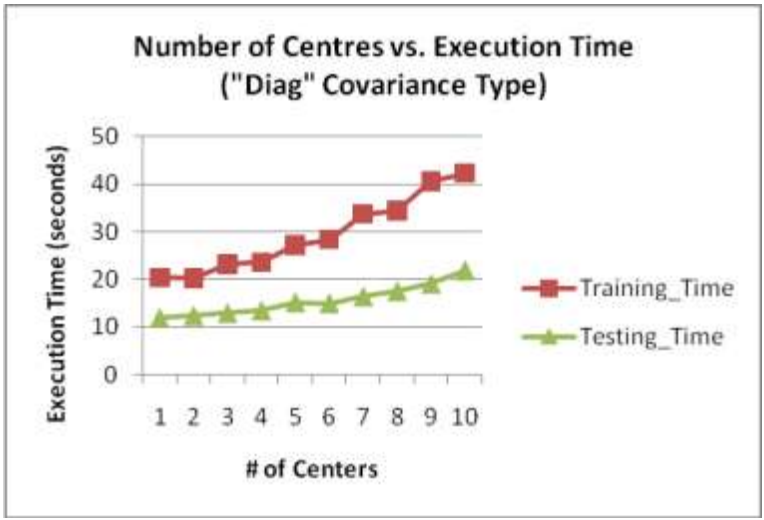

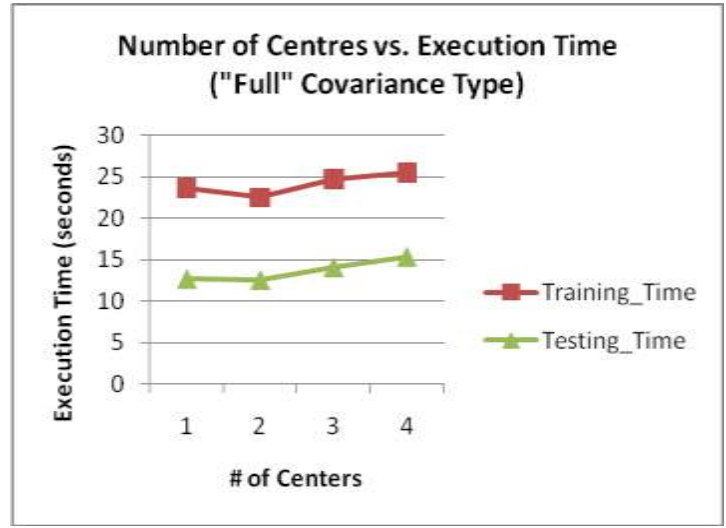

Figure 3.   Relationship between the number of centers and execution time for GMM "diag" and "full" covariance types.

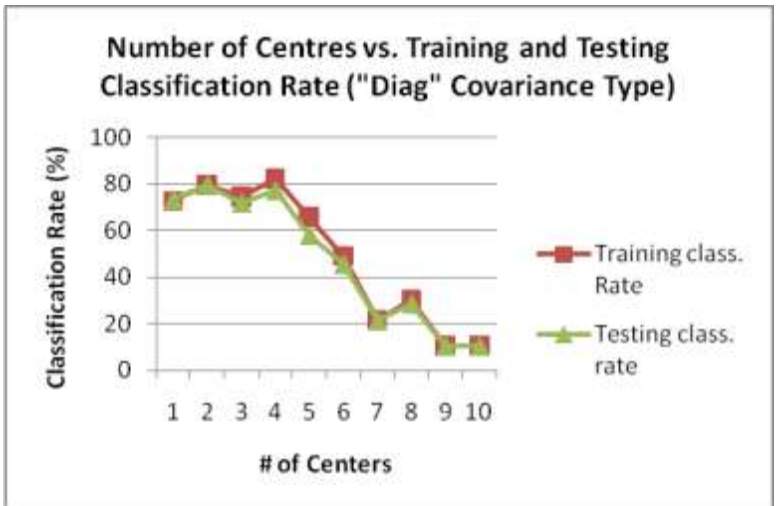

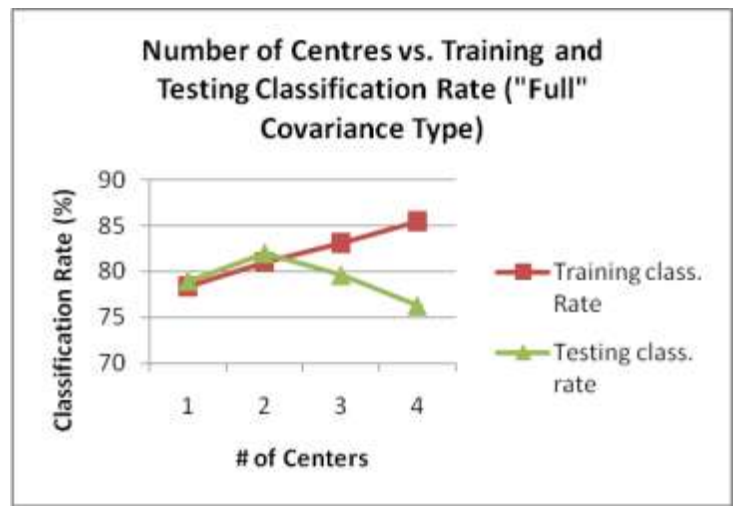

Figure 4.   Relationship between the number of centers and recognition rate for GMM "diag" and "full" covariance types.

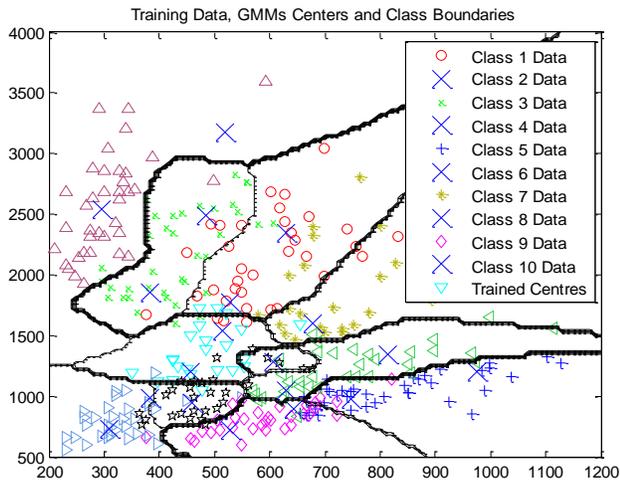

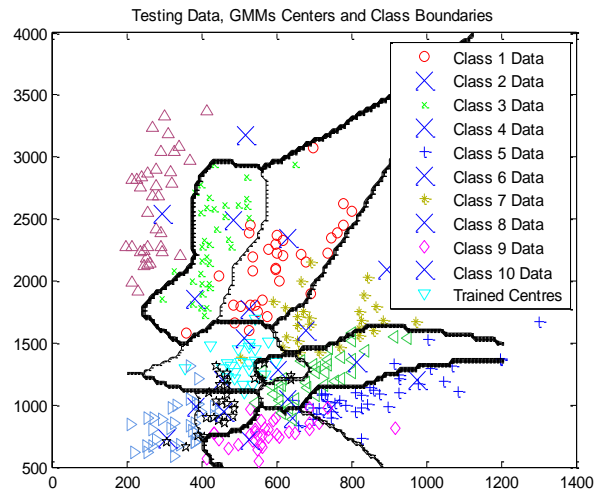

Figure 5.   Class boundaries generated by the GMM Model for training and testing.

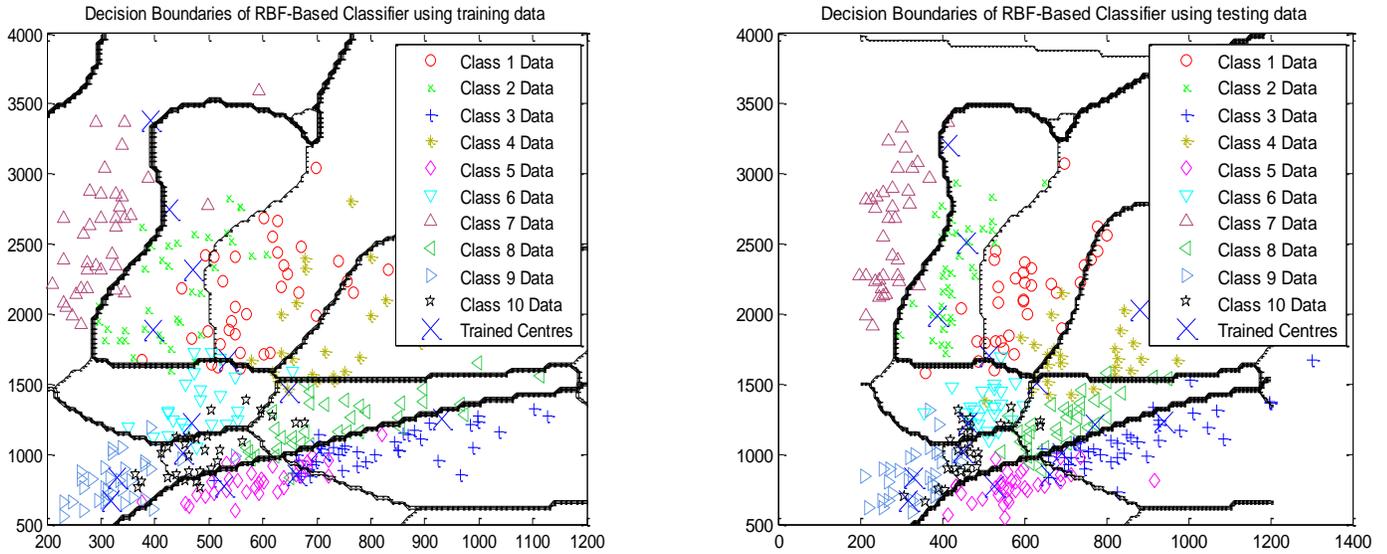

Figure 6.   Decision boundaries of the RBF-based classifier using training and testing data.

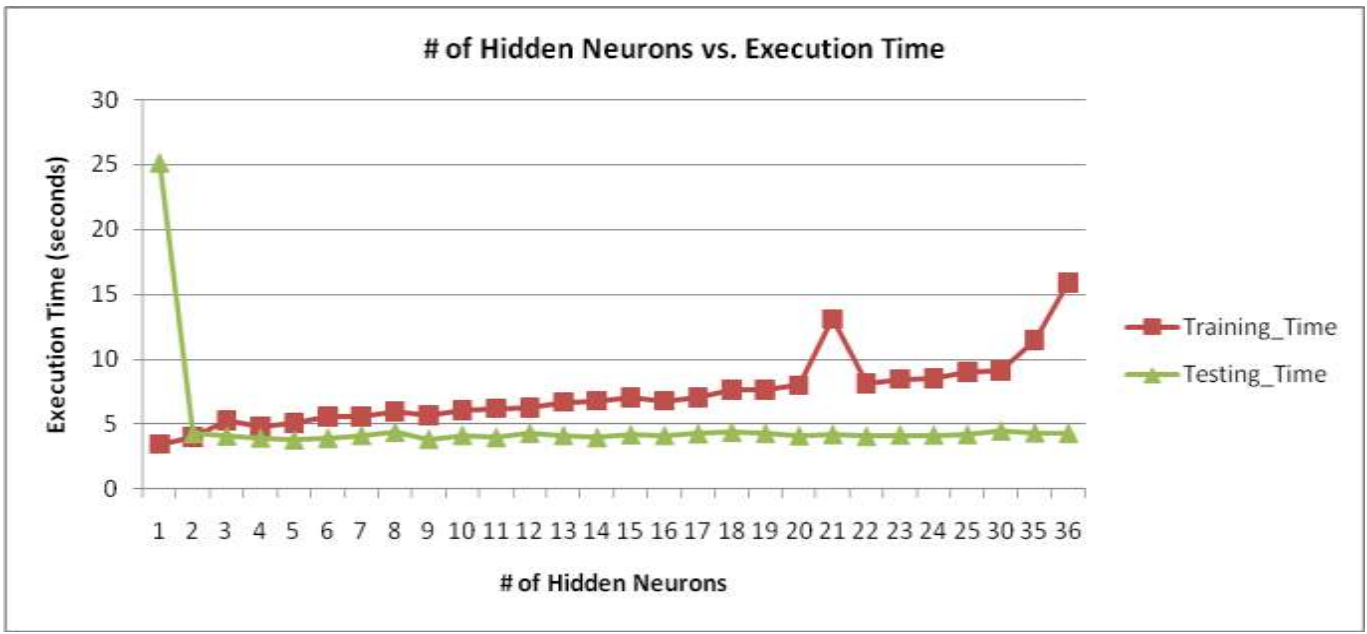

Figure 7.   Relationship between the number of hidden neurons and the execution time.

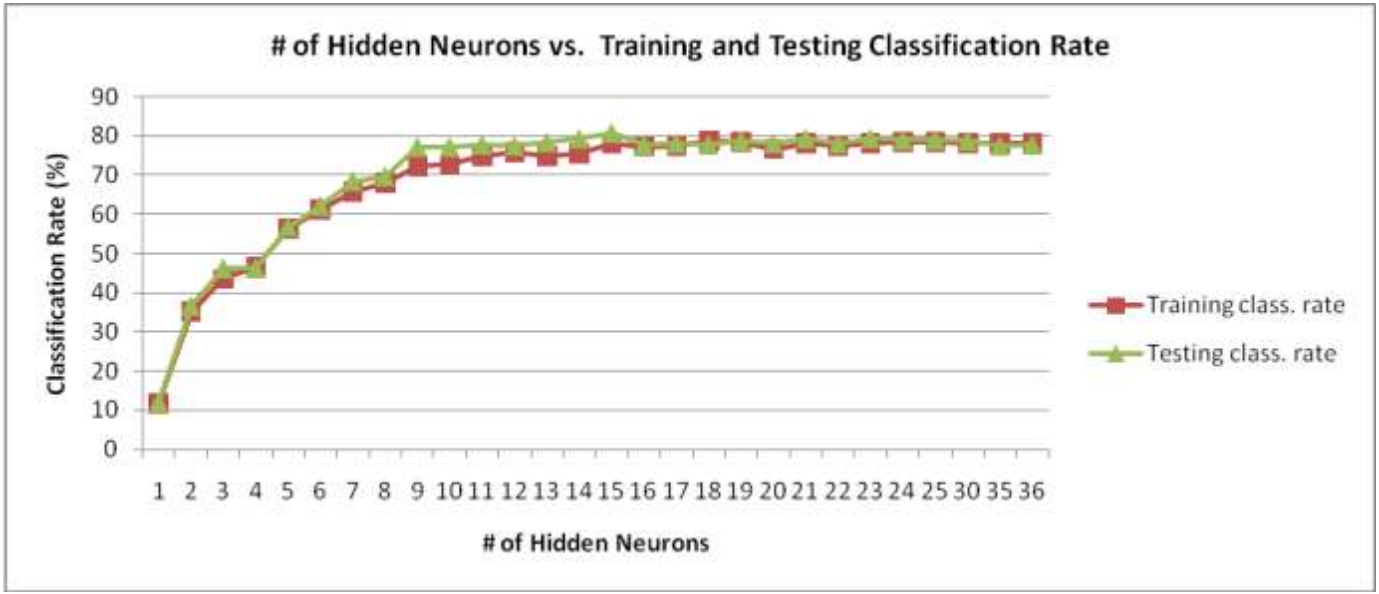

Figure 8.   Relationship between the number of hidden neurons and recognition rate.

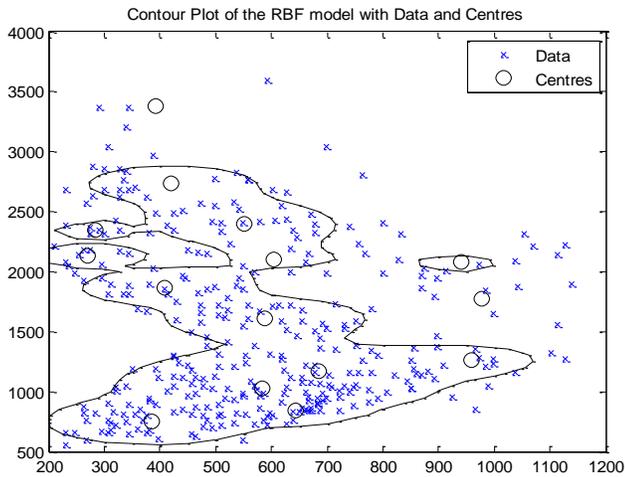

Figure 9.   Contour plot of the RBF model showing the 15 hidden neurons.

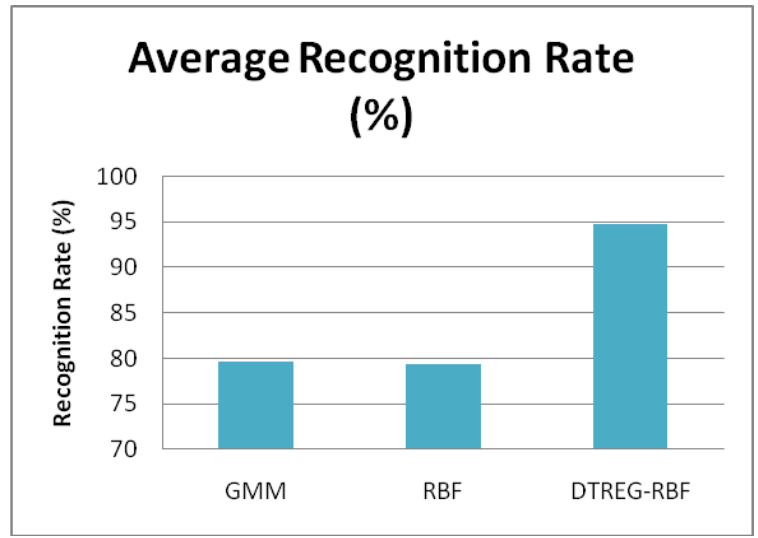

Figure 10.  A comparison of GMM, RBF and DTREG RBF models by recognition rate.